%% file: CVPR2025-FDConv.tex
\documentclass[10pt,twocolumn,letterpaper]{article}
\input{preamble}

\usepackage{cvpr}              

\usepackage{times}
\usepackage{epsfig}
\usepackage{graphicx}
\usepackage{amsmath}
\usepackage{amssymb}
\usepackage{booktabs}
\usepackage{cite}
\usepackage{color}
\usepackage{times}
\usepackage{epsfig}
\usepackage{graphicx}
\usepackage{graphics}
\usepackage{amsmath}
\usepackage{amssymb}
\usepackage{multirow}
\usepackage{mathrsfs}
\usepackage{rotating}
\usepackage{bbm}
\usepackage{verbatim}
\usepackage{array}
\usepackage{tabularx}
\usepackage{url}
\usepackage{marvosym}
\usepackage{xcolor}
\definecolor{myblue}{RGB}{0,0,128}
\definecolor{mygray}{RGB}{128, 128, 128}
\definecolor{cvprblue}{rgb}{0.21,0.49,0.74}
\usepackage{colortbl}
\usepackage{algorithm}
\usepackage{algorithmicx}
\usepackage{algpseudocode}
\PassOptionsToPackage{table}{xcolor}
\usepackage[table]{xcolor}
\usepackage[pagebackref,breaklinks,colorlinks,citecolor=cvprblue]{hyperref}
\usepackage[capitalize]{cleveref}
\crefname{section}{Sec.}{Secs.}
\Crefname{section}{Section}{Sections}
\Crefname{table}{Table}{Tables}
\crefname{table}{Tab.}{Tabs.}

\def\paperID{2054} 
\def\confName{CVPR}
\def\confYear{2025}

\begin{document}

\title{
Frequency Dynamic Convolution for Dense Image Prediction
}

\author{
Linwei Chen$^1$ \qquad Lin Gu$^{2,3}$ \qquad Liang Li$^{4}$ \qquad Chenggang Yan$^{5,6}$ \qquad Ying Fu$^{1*}$ \\
$^1$Beijing Institute of Technology 
\qquad 
$^2$RIKEN \qquad 
$^3$The University of Tokyo  \\
$^4$Chinese Academy of Sciences \quad 
$^5$Hangzhou Dianzi University \quad
$^6$Tsinghua University
\\
{\tt\footnotesize
chenlinwei@bit.edu.cn;
lin.gu@riken.jp;
liang.li@ict.ac.cn;
cgyan@hdu.edu.cn; 
fuying@bit.edu.cn}
}
\maketitle
\let\thefootnote\relax\footnotetext{*Corresponding Author}
\begin{abstract}

While Dynamic Convolution (DY-Conv) has shown promising performance by enabling adaptive weight selection through multiple parallel weights combined with an attention mechanism, the frequency response of these weights tends to exhibit high similarity, resulting in high parameter costs but limited adaptability.
In this work, we introduce Frequency Dynamic Convolution (FDConv), a novel approach that mitigates these limitations by learning a fixed parameter budget in the Fourier domain. FDConv divides this budget into frequency-based groups with disjoint Fourier indices, enabling the construction of frequency-diverse weights without increasing the parameter cost. 
To further enhance adaptability, we propose Kernel Spatial Modulation (KSM) and Frequency Band Modulation (FBM). KSM dynamically adjusts the frequency response of each filter at the spatial level, while FBM decomposes weights into distinct frequency bands in the frequency domain and modulates them dynamically based on local content.
Extensive experiments on object detection, segmentation, and classification validate the effectiveness of FDConv. We demonstrate that when applied to ResNet-50, FDConv achieves superior performance with a modest increase of +3.6M parameters, outperforming previous methods that require substantial increases in parameter budgets (e.g., CondConv +90M, KW +76.5M). Moreover, FDConv seamlessly integrates into a variety of architectures, including ConvNeXt, Swin-Transformer,  offering a flexible and efficient solution for modern vision tasks. 
The code is made publicly available at \href{https://github.com/Linwei-Chen/FDConv}{https://github.com/Linwei-Chen/FDConv}.

\end{abstract}



\begin{figure}[t!]
\centering
\scalebox{1.0}{
\begin{tabular}{cc}
\hspace{-3.918mm}
\includegraphics[width=1.0\linewidth]{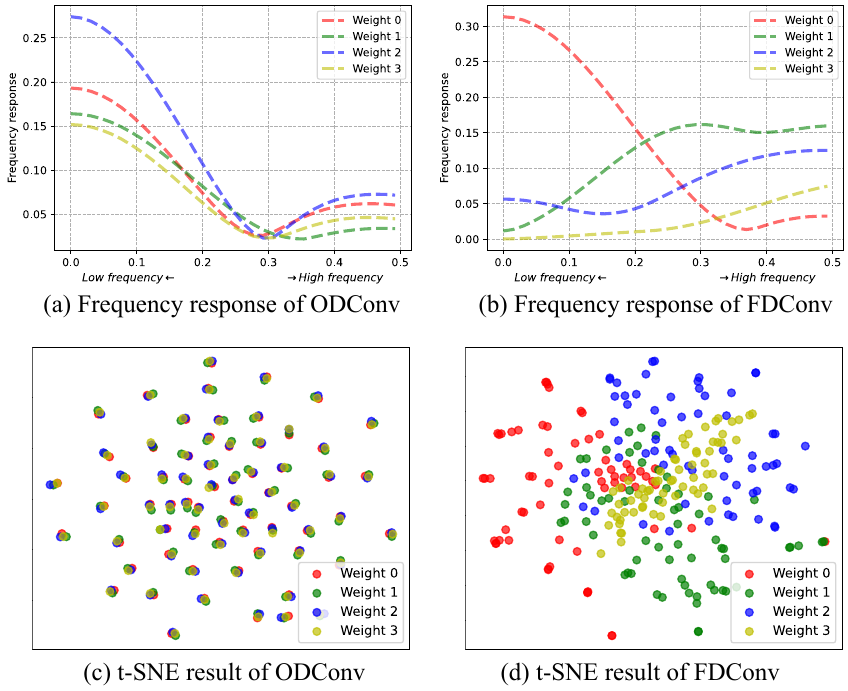} \\
\vspace{-6mm}
\end{tabular}}
\caption{
Weight frequency responses and t-SNE analyses. We set the number of weights to 4 to align with ODConv~\cite{2022odconv}.
(a) The frequency responses of the four parallel weights in ODConv are highly similar, indicating limited diversity. 
(b) In contrast, FDConv shows distinct frequency responses for each weight, spanning different parts of the frequency spectrum. 
(c) The t-SNE plot for ODConv reveals that the filters in the four weights are closely clustered, suggesting a lack of diversity. 
(d) The t-SNE plot for FDConv shows that the filters in the four weights have different distributions, indicating greater diversity.
}
\label{fig:intro}
\vspace{-5mm} 
\end{figure}

\section{Introduction}
\label{sec:intro}
Convolution, the core operation in ConvNets, has driven decades of advancement in computer vision~\cite{resnet2016, 2022convnet, zou2024eventhdr, liu2024siamese, lai2024hyperspectral, zhang2024deep, tian2023transformer, fu2023category, li2024supervise, 2023lis, 2021efficienthybrid, 2022hybridsupervised, chen2022consistency, chen2024semantic, guan2023hrpose, li2024yolo}. Essential for capturing local patterns and building hierarchical representations, it remains fundamental in modern architectures~\cite{2022convnet, 2024dcnv4, 2024unireplknet, 2021deit}.


Building upon the success of standard convolution, Dynamic Convolution (DY-Conv)~\cite{2019condconv, 2020dyconv} offers a more adaptive and efficient approach. 
Unlike standard convolution with fixed weights, DY-Conv uses multiple parallel weights combined by an attention module, allowing sample-specific weight adaptation with minimal extra computation.

However, our analysis in Figure~\ref{fig:intro} reveals that traditional dynamic convolution~\cite{2019condconv, 2020dyconv, 2022odconv, 2024kw} \textit{lack of frequency responses diversity} in their parallel weights.
As shown in Figure~\ref{fig:intro}{\color{red}(a)}, these weights exhibit highly similar frequency responses, while the t-SNE visualization in Figure~\ref{fig:intro}{\color{red}(c)} indicates that filters within ODConv~\cite{2022odconv} are clustered closely.
Despite a significant increase in parameters (\eg, $4\times$ in~\cite{2019condconv, 2020dyconv, 2022odconv, 2024kw}), this limited frequency diversity reduces the model’s ability to adaptively capture frequency information. 
For example, extracting low-frequency components helps suppress noise~\cite{2023lis}, while high-frequency components capture details and boundaries~\cite{2024freqfusion, 2021dynamic, exfuse2018, 2023sfnet}, which are vital for foreground-background differentiation.

\begin{figure*}[t!]
\centering
\scalebox{0.98}{
\begin{tabular}{cc}
\includegraphics[width=0.98\linewidth]{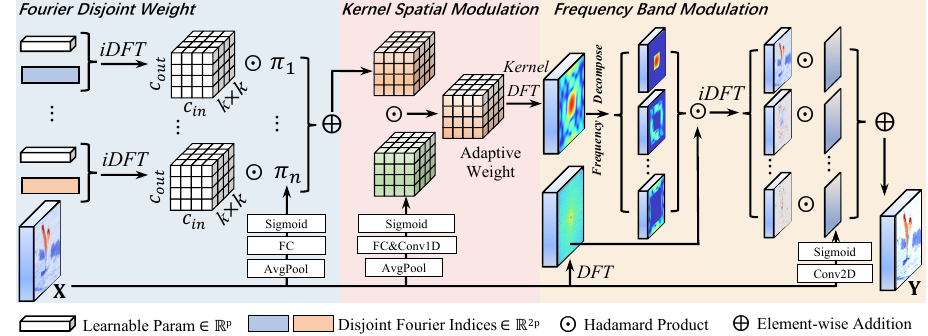} \\
\end{tabular}}
\caption{
Illustration of the proposed Frequency Dynamic Convolution, which consists of the Fourier Disjoint Weight (FDW), Kernel Spatial Modulation (KSM), and Frequency Band Modulation (FBM) modules.
FC indicates fully connected layer.
}
\label{fig:method}
\vspace{-2.18mm} 
\end{figure*}

To address these limitations, we propose Frequency Dynamic Convolution (FDConv), as shown in Figure~\ref{fig:method}. 
It is designed to enhance frequency adaptability without incurring excessive parameter overhead. Our approach is based on three core modules: Fourier Disjoint Weight, Kernel Spatial Modulation, and Frequency Band Modulation.


Unlike traditional methods~\cite{2019condconv, 2020dyconv, 2022odconv}, which learn weights in the spatial domain, the Fourier Disjoint Weight (FDW) constructs kernel weights by learning spectral coefficients in the Fourier domain. These coefficients are divided into frequency-based groups, each with a disjoint set of Fourier indices. An inverse Discrete Fourier Transform (iDFT) is then applied to these groups, converting them into spatial weights. This disjoint grouping enables each weight to exhibit distinct frequency responses (as shown in Figure~\ref{fig:intro}{\color{red}(b)}), ensuring high diversity among the learned weights (also shown in Figure~\ref{fig:intro}{\color{red}(d)}).

Kernel Spatial Modulation (KSM) enhances flexibility by precisely adjusting the frequency response of each filter at the spatial level within the kernel. By combining local and global channel information, KSM generates a dense matrix of modulation values that finely tunes each individual weight element. 
This fine-grained control enables FDConv to dynamically adapt each filter element, allowing for frequency response adjustment across the entire kernel.

Frequency Band Modulation (FBM) decomposes weights into distinct frequency bands in the frequency domain, enabling spatially variant frequency modulation. 
It allows each frequency band of the weight to be adjusted independently across spatial locations. 
Unlike traditional dynamic convolutions, which apply fixed frequency responses across spatial dimensions, FBM decomposes weights into distinct frequency bands and dynamically modulates them based on local content. 
This design enables the model to selectively emphasize or suppress frequency bands across different regions, adaptively capturing diverse frequency information in a spatially variant manner.

Moreover, unlike previous works~\cite{2019condconv, 2020dyconv, 2022odconv}, which increase parameter costs by a factor of $n$ (where $n$ is the number of weights, typically $n < 10$~\cite{2019condconv, 2020dyconv, 2022odconv}), our FDConv maintains a fixed parameter budget while generating a large number of frequency-diverse weight kernels ($n > 10$) by dividing parameters in the Fourier domain into disjoint frequency-based groups. This design allows the model to efficiently learn weights with distinct frequency responses without burdening parameter cost.

Extensive experiments on object detection, instance segmentation, semantic segmentation, and image classification validate the effectiveness of FDConv. 
For example, when applied to ResNet-50, FDConv achieves superior performance with a modest increase of +3.6M parameters, outperforming previous methods that require substantial increases in parameter budgets (\eg, CondConv +90M, DY-Conv +75.3M, ODConv +65.1M, KW +76.5M)~\cite{2019condconv, 2020dyconv, 2022odconv, 2024kw}. FDConv can be seamlessly integrated into various architectures, including ConvNeXt and Swin Transformer, where it replaces the linear layer (as a 1$\times$1 convolution), offering a versatile and efficient solution.

\begin{itemize}
\item We conduct a comprehensive exploration of dynamic convolution using frequency analysis. Our findings reveal that the parameters of traditional dynamic convolution methods exhibit high homogeneity in frequency response across learned parallel weights, resulting in high parameter redundancy and limited adaptability.
\item We introduce the Fourier Disjoint Weight (FDW), Kernel Spatial Modulation (KSM), and Frequency Band Modulation (FBM) strategies. FDW constructs multiple weights with diversified frequency responses without increasing the parameter cost, KSM enhances the representation power by adjusting weights element-wise, and FBM improves convolution by precisely extracting frequency bands in a spatially variant manner.
\item We demonstrate that our approach can be easily integrated into existing ConvNets and vision transformers. Comprehensive experiments on segmentation tasks show that it surpasses previous state-of-the-art dynamic convolution methods, requiring only a minor increase in parameters, consistently demonstrating its effectiveness.
\end{itemize}

\section{Related Work}
\noindent\textbf{Feature Recalibration.} 
Feature recalibration through attention mechanisms has proven highly effective in deep learning models. Methods such as RAN~\cite{2017ran}, SE~\cite{hu2018squeeze}, CBAM~\cite{2018cbam}, GE~\cite{2018ge}, SRM~\cite{2019srm}, ECA~\cite{2020eca}, and SimAtt~\cite{2021simam} focus on adaptively emphasizing informative features or suppressing irrelevant ones across channels and spatial dimensions of the feature map, \ie, channel and spatial attention. In contrast, our approach introduces frequency-specific recalibration for convolution weights.

\vspace{+0.518mm}
\noindent\textbf{Dynamic Weight Networks.}
Recently, dynamic networks have shown to be effective in various computer vision tasks.
Dynamic Filter Networks \cite{2016dynamic} and Kernel Prediction Networks \cite{2018kpn} generate sample-adaptive filters conditioned on the input. In contrast, Hypernetworks~\cite{2016hypernetworks} generate weights for a larger recurrent network instead of ConvNets. Building upon similar idea, CARAFE~\cite{carafe} and Involution \cite{2021involution} have developed efficient modules that predict spatially variant convolution weights.

Dynamic convolution methods~\cite{2019condconv, 2020dyconv} learn multiple parallel weights and adaptively mix them linearly using attention modules. CondConv~\cite{2019condconv} uses a sigmoid function for weight fusion, while DY-Conv improves upon this by using a softmax function~\cite{2020dyconv}. 
Inspired by SE~\cite{hu2018squeeze}, WeightNet~\cite{2020weightnet}, CGC~\cite{2020cgc}, and WE~\cite{2020we}, these methods design various attention modules to adjust convolutional weights in ConvNets. ODConv~\cite{2022odconv} further enhances the attention module by predicting channel-wise, filter-wise, and spatial-wise attention values to adjust the weights.

To mitigate the increased parameter overhead of multiple weights, methods like DCD~\cite{2021DCD} and PEDConv~\cite{2021PEConv} use matrix decomposition techniques to construct low-rank weight matrices, reducing computational complexity. More recently, KW~\cite{2024kw} introduced a decomposition approach where kernel weights are divided into smaller, shareable units across different stages and layers, enabling dynamic kernel reconstruction with fewer parameters. 

In contrast, our FDConv addresses the heavy parameter cost and limited diversity of weights from the frequency aspect, offering a new solution.

\vspace{+0.518mm}
\noindent{\bf Frequency Domain Learning.}
Frequency-domain analysis has long been a cornerstone of signal processing~\cite{2009digital, 2000digital}. 
Recently, these techniques have been leveraged in deep learning, influencing model optimization strategies~\cite{2019fourier}, robustness~\cite{2023improving}, and generalization abilities~\cite{2020highfrequency} in Deep Neural Networks (DNNs). 
Moreover, the integration of frequency-domain methods into DNNs has proven effective for learning global features~\cite{2020ffc, 2021gfnet, 2021FNO, 2022AFNO, 2023adaptivefrequency} and enhancing domain-generalizable representations~\cite{2023dff}. 
FcaNet~\cite{2021fcanet} demonstrates that frequency information benefits feature recalibration, while FreqFusion~\cite{2024freqfusion} shows its advantages in feature fusion. Some studies~\cite{2022flc, chen2024semantic, 2023fixasap} have improved downsampling operations by addressing high-frequency components that lead to aliasing. FADC~\cite{2024fadc} enhances dilated convolution by adjusting dilation based on the frequency characteristics of the features. Our method incorporates a frequency-based perspective into dynamic convolution, improving its ability to learn diversified weights for capturing a wider range of frequency information.


\section{Method}
An overview of the proposed Frequency Dynamic Convolution (FDConv) framework is shown in Figure~\ref{fig:method}. 
This section first introduces the concept of Fourier Disjoint Weights, followed by a detailed exploration of two key strategies: Kernel Spatial Modulation and Frequency Band Modulation, which are designed to fully leverage the frequency adaptability of FDConv in the kernel spatial and frequency domains, respectively.

\subsection{Fourier Disjoint Weight}
\textbf{Dynamic Convolution.}
For a standard convolutional layer, it can be formulated as $\mathbf{Y} = \mathbf{W} * \mathbf{X}$, where $\mathbf{X} \in \mathbb{R}^{h \times w \times C_{\text{in}}}$ and $\mathbf{Y} \in \mathbb{R}^{h \times w \times C_{\text{out}}}$ are the input and output features, respectively. 
Here, $C_{\text{in}}$ and $C_{\text{out}}$ represent the number of input and output feature channels, and $h \times w$ denotes the spatial size. 
The weight $\mathbf{W} \in \mathbb{R}^{k \times k \times C_{\text{in}} \times C_{\text{out}}}$ consists of $C_{\text{out}}$ convolutional filters, each with a spatial size of $k \times k$.

Dynamic convolution~\cite{2020dyconv, 2019condconv} enhances the adaptability of convolutional layers by replacing the static weight $\mathbf{W}$ in standard convolution with a combination of $n$ distinct weights $\{\mathbf{W}_1, \ldots, \mathbf{W}_n\}$, each of the same dimension. 
The contribution of each kernel is modulated by a set of attention-based coefficients $\{\pi_1, \ldots, \pi_n\}$, which are dynamically generated. 
Typically, these coefficients are derived by applying global average pooling on the input, followed by a fully connected (FC) layer. 
This dynamic convolution operation can be expressed as

\begin{equation}
\label{eq:dyconv}
\begin{aligned}
\mathbf{W} = \pi_1 \mathbf{W}_1 + \ldots + \pi_n \mathbf{W}_n.
\end{aligned}
\end{equation}

Despite the increased parameter cost by a factor of $n$, we expect dynamic convolution to learn diverse weights. However, our analysis reveals, as shown in Figure~\ref{fig:intro}, that the frequency responses of parallel weights are highly similar. This lack of frequency diversity limits the model's ability to adaptively capture features across different frequency bands, reducing the flexibility of dynamic convolution.

\begin{figure}[t!]
\centering
\scalebox{1.0}{
\begin{tabular}{cc}
\hspace{-2.918mm}
\includegraphics[width=0.99918\linewidth]{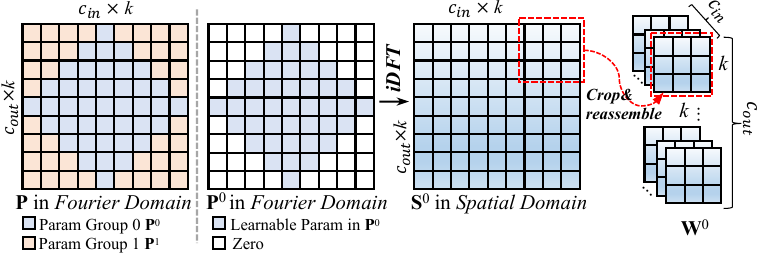} \\
\vspace{-6mm}
\end{tabular}}
\caption{
Illustration of Fourier Disjoint Weight (FDW). 
The left figure illustrates the division of parameters into disjoint groups, ranging from low frequencies (center) to high frequencies (border). In this example, $n = 2$ groups are shown.
The right figure demonstrates how to obtain the convolution weights from the learnable parameter group 0. It first transforms the learnable parameters with specific Fourier indices (with all other Fourier indices set to zero) using the inverse Discrete Fourier Transform (iDFT). 
The resulting spatial weights are then obtained by cropping the iDFT result into $k \times k$ patches and reshaping them into a weight tensor of size $k \times k \times C_{\text{in}} \times C_{\text{out}}$.
}
\label{fig:FDW}
\vspace{-5mm} 
\end{figure}

\vspace{+0.518mm}
\noindent\textbf{Overview of Fourier Disjoint Weight.}  
To construct multiple parallel weights with high frequency response diversity without increasing parameter costs, we propose Fourier Disjoint Weight (FDW). 
Unlike previous methods~\cite{2020dyconv, 2019condconv, 2022odconv}, which are limited to a small number of kernels ($n < 10$) due to high parameter costs, FDW can generate $n > 10$ diversified weights. 

The core concept of FDW is learning spectral coefficients in the Fourier domain with disjoint sets of Fourier indices, rather than in the traditional spatial domain. FDW involves three steps to construct $n$ weights:  
1) \textit{Fourier disjoint grouping.} Divide a fixed number of parameters into $n$ groups with disjoint Fourier indices.
2) \textit{Fourier to spatial transformation.} Convert each group of parameters from the Fourier domain to the spatial domain using the Inverse Discrete Fourier Transform (iDFT).  
3) \textit{Reassembling.} Crop the transformed results in the spatial domain into $k \times k$ patches and reassemble them into the standard weight shape of $k \times k \times C_{\text{in}} \times C_{\text{out}}$.

\vspace{+0.518mm}
\noindent\textbf{Fourier Disjoint Grouping.}  
Given a parameter budget of $k \times k \times C_{\text{in}} \times C_{\text{out}}$, FDW first treats these parameters as learnable spectral coefficients in the Fourier domain, reshaping them into $\mathbf{P} \in \mathbb{R}^{kC_{\text{in}} \times kC_{\text{out}}}$.
Each parameter is associated with a Fourier index $(u, v)$, \ie, coordinates in the Fourier domain that indicate frequency.  
FDW then sorts these parameters from low to high frequency based on the $L_2$ norm of the Fourier index, $|| (u, v) ||_2$, and divides them uniformly into disjoint $n$ set, $\{\mathbf{P}^0, \dots, \mathbf{P}^{n-1}\}$.

As shown on the left side of Figure~\ref{fig:FDW}, we divide the learnable parameters into $n=2$ groups for simplicity in the demonstration, where the center represents low frequencies and the border represents high frequencies. Moreover, the number of groups, $n$, can be set to a large value ($n > 10$), allowing for the generation of a large number of diversified weights without increasing the parameter cost.

\vspace{+0.518mm}
\noindent\textbf{Fourier to Spatial Transformation.}  
To obtain the weights, FDW transforms each group of parameters into the spatial domain using the inverse Discrete Fourier Transform (iDFT). This can be formulated as:
\begin{equation}
\begin{aligned}
\mathbf{S}^{i}_{p,q} = \sum_{u=0}^{kC_{\text{in}}-1} \sum_{v=0}^{kC_{\text{out}}-1} \mathbf{P}^{i}_{u,v} e^{i 2\pi \left( \frac{p}{kC_{\text{in}}} u + \frac{q}{kC_{\text{out}}} v \right)} 
\end{aligned}
\vspace{-2.18mm}
\end{equation}
where $\mathbf{P}^{i}_{u,v}$ is the parameter with Fourier index $(u, v)$ in the $i$-th group $\mathbf{P}^{i}$.  
As shown on the right side of Figure~\ref{fig:FDW}, if the Fourier index $(u, v)$ is assigned to the $i$-th group, then $\mathbf{P}^{i}_{u,v} = \mathbf{P}_{u,v}$, otherwise, $\mathbf{P}^{i}_{u,v} = 0$.  
$\mathbf{S}^{i}_{p,q} \in \mathbb{R}^{kC_{\text{in}} \times kC_{\text{out}}}$ represents the element at position $(p, q)$ in the converted results in the spatial domain.

\vspace{+0.518mm}
\noindent \textbf{Reassembling.}  
As shown on the right side of Figure~\ref{fig:FDW}, the final $i$-th weight $\mathbf{W}^i \in \mathbb{R}^{k \times k \times C_{\text{in}} \times C_{\text{out}}}$ can be obtained by cropping $\mathbf{S}^{i} \in \mathbb{R}^{kC_{\text{in}} \times kC_{\text{out}}}$ into $C_{\text{in}} \times C_{\text{out}}$ patches of size $k \times k$ and reassembling them to form $\mathbf{W}^i$.

Since the parameters are divided according to frequency, $\mathbf{S}^{i}$ contains only the frequency components of a specific band. Therefore, each weight $\mathbf{W}^i$ derived from $\mathbf{S}^{i}$ exhibits a distinct frequency response compared to $\mathbf{W}^j$ when $i \neq j$.  
This ensures that the frequency responses of the constructed weights are diversified. After a linear mixture, as described in Equation~\eqref{eq:dyconv}, FDW can adaptively adjust the frequency response of the combined weight based on the input sample.

Note that FDW can also be applied to linear layers in modern vision architectures, such as Transformers~\cite{2020vit, 2021swin}, which are equivalent to convolutions with a kernel size of 1.

\begin{figure}[t!]
\centering
\scalebox{0.9918}{
\begin{tabular}{cc}
\hspace{-2.18mm}
\includegraphics[width=0.98\linewidth]{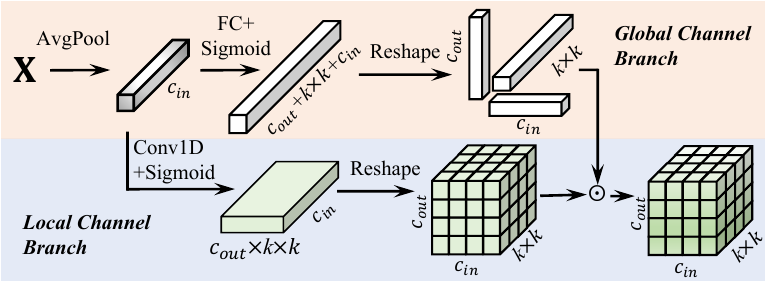} \\
\vspace{-6mm}
\end{tabular}}
\caption{
Illustration of Kernel Spatial Modulation (KSM).  
The KSM consists of two branches: the global channel branch and the local channel branch.   
The local channel branch employs a very lightweight 1-D convolution to obtain local channel information and predicts a dense modulation matrix of size $k \times k \times C_{\text{in}} \times C_{\text{out}}$.  
The global branch uses a fully connected layer to obtain the global channel information and predicts three dimension-wise modulation values along the input channel, output channel, and kernel spatial dimensions. 
The two branches are fused to obtain the final weight modulation matrix.
}
\label{fig:KSM}
\vspace{-2.18mm} 
\end{figure}

\subsection{Kernel Spatial Modulation}  
By ensuring the diversity of frequency responses of parallel weights in the Fourier domain, FDW can adaptively adjust the frequency response of the combined weight based on the input sample after a linear mixture, as described in Equation~\eqref{eq:dyconv}. 
However, this weight-wise mixture is too coarse and cannot independently adjust the frequency response of each $k \times k$ filter in the weight.

To address this issue, we propose Kernel Spatial Modulation (KSM), which predicts a dense modulation matrix $\mathbf{\alpha} \in \mathbb{R}^{k \times k \times C_{\text{in}} \times C_{\text{out}}}$, instead of a sparse vector. As shown in Figure~\ref{fig:KSM}, KSM consists of a local channel branch and a global channel branch. 

\vspace{+0.518mm}
\noindent \textbf{Local Channel Branch.}  
While global fully connected layers are commonly used for modulation value prediction~\cite{hu2018squeeze, 2019condconv, 2020dyconv, 2022odconv}, they are not suitable for predicting dense modulation matrices due to their large parameter and computational costs.
To address this, the local channel branch employs a lightweight 1-D convolution, which has been proven to be efficient and effective~\cite{2020eca}.
It captures local channel information and predicts a dense modulation matrix of size $k \times k \times C_{\text{in}} \times C_{\text{out}}$. This approach significantly reduces parameters and computational complexity while maintaining the ability to learn fine-grained modulation for each element in the weight.

\vspace{+0.518mm}
\noindent \textbf{Global Channel Branch.}  
Though the local channel branch can efficiently predict the modulation matrices, it lacks global information, which is crucial for weight adjustments. 
To complement the local channel branch, the global channel branch uses a fully connected layer to capture global channel information and predict a sparse modulation vector for efficiency. 
Specifically, it predicts three dimension-wise modulation values: one for the input channel, one for the output channel, and one for the kernel spatial dimensions, ensuring that both local and global contextual information are incorporated for adaptive modulation.

In this way, the proposed KSM is able to leverage both local and global information, enabling more precise and context-aware modulation of each filter in the weights.

\subsection{Frequency Band Modulation}
While the proposed Fourier Disjoint Weight (FDW) and Kernel Spatial Modulation (KSM) modules substantially enhance adaptability by ensuring frequency diversity and element-wise adjustments, they remain spatially invariant, as is typical in dynamic convolution~\cite{2019condconv, 2020dyconv, 2022odconv, 2024kw}. 
This spatial invariance, where weights are shared across the entire feature map, restricts convolutional layers from dynamically adapting frequency responses to spatially varying content, limiting their ability to fully capture complex structures across the image.

Natural images and their corresponding features exhibit large spatial variation, which necessitates frequency-specific adaptations for optimal feature extraction. For example, extracting low-frequency components is vital for suppressing feature noise~\cite{2023lis}, while high-frequency components are essential for capturing fine details and boundaries~\cite{2024freqfusion, 2021dynamic}, which are crucial for distinguishing the foreground from the background.

\vspace{+0.518mm}
\noindent\textbf{Overview of Frequency Band Modulation.}
To address the need for spatially dynamic frequency modulation, we propose Frequency Band Modulation (FBM).
FBM decomposes the convolutional kernel into multiple frequency bands in the frequency domain and applies spatially specific modulations, adaptively adjusting each frequency component across different spatial locations.

The Frequency Band Modulation operates in the following key steps:
1) Kernel frequency decomposition. Decomposing the frequency response of the convolution weight into different frequency bands.
2) Convolution in the Fourier domain. Performing convolution in the Fourier domain.
3) Spatially variant modulation. Predicting modulation values for each frequency band of the convolution weight across different spatial locations.

The core formulation of FBM is given by:
\vspace{-2mm}
\begin{equation}
\begin{aligned}
\mathbf{Y} = \sum_{b=0}^{B-1} (\mathbf{A}_b \odot (\mathbf{W}_b * \mathbf{X})),
\end{aligned}
\vspace{-2mm}
\end{equation}

where $\mathbf{X}$ and $\mathbf{Y} \in \mathbb{R}^{h \times w}$ are the input and output feature maps. Note that we omit the channel dimension for simplicity.
$\mathbf{A}_b \in \mathbb{R}^{h \times w}$ representing spatial modulation values specific to the $b$-th frequency band, 
And $\mathbf{W}_b$ the $b$-th frequency band of weight. 
FBM enables adjusting the frequency responses for each spatial location of feature map.

\vspace{+0.518mm}
\noindent\textbf{Kernel frequency decomposition.}
To decompose the convolution kernel into distinct frequency bands, FBM first pads the kernel $\mathbf{W}$ to match the feature map size~\cite{2014fasttrainingfft}, and then applies a set of binary masks $\mathcal{M}_b$ to isolate specific frequency ranges:
\begin{equation}
\mathbf{W}_{b} = \mathcal{F}^{-1}(\mathcal{M}_{b} \odot \mathcal{F}(\mathbf{W})),
\end{equation}
where $\mathcal{F}$ and $\mathcal{F}^{-1}$ denote the DFT and inverse DFT, and $\mathcal{M}_{b}$ is a binary mask isolating specific frequency ranges:
\begin{equation}
\mathcal{M}_{b}(u, v) =
\begin{cases} 
1 & \text{if } \psi_b \leq \max(|u|,|v|) < \psi_{b+1} \\
0 & \text{otherwise}
\end{cases}
\end{equation}
Here, $\psi_b$ and $\psi_{b+1}$ are thresholds from the predefined frequency set $\{0, \psi_1, \dots, \psi_{B-1}, \frac{1}{2}\}$, where $(u, v)$ denote the horizontal and vertical frequency indices. 
By default, we decompose the frequency spectrum into four distinct bands using an octave-based partitioning strategy~\cite{2024spatialfrequency}. 
The thresholds for dividing the frequency bands are $\{0, \frac{1}{16}, \frac{1}{8}, \frac{1}{4}, \frac{1}{2}\}$.

\vspace{+0.518mm}
\noindent\textbf{Convolution in the Fourier Domain.} 
After obtaining the frequency-specific weights $\mathbf{W}_b$, the output for the corresponding frequency band can be computed as follows:
\begin{equation}
\begin{aligned}
\mathbf{Y}_b = \mathbf{W}_b * \mathbf{X},
\end{aligned}
\end{equation}
where $\mathbf{Y}_b$ represents the output for the $b$-th frequency band.

However, as discussed in~\cite{2022flc, 2009digital}, obtaining specific frequency bands of $\mathbf{W}$ directly in the spatial domain is challenging. For instance, $\mathbf{W}_b$ would need to be infinitely large to isolate the low-frequency part of $\mathbf{W}$, since the ideal low-pass filter $sinc$ has infinite support in spatial domain~\cite{2009digital}. 

To overcome this limitation, we perform the convolution in the Fourier domain rather than in the spatial domain. 
According to the Convolution Theorem~\cite{2009digital}, convolution in the spatial domain is equivalent to pointwise multiplication of the Fourier transforms in the frequency domain. Therefore, we formulate the frequency-specific convolution as:
\begin{equation}
\begin{aligned}
\label{eq:y_b}
\mathbf{Y}_b = \mathcal{F}^{-1}\left(\left(\mathcal{M}_{b} \odot \mathcal{F}(\mathbf{W})\right) \odot \mathcal{F}(\mathbf{X})\right).
\end{aligned}
\end{equation}
This formulation enables the efficient computation of convolutions for each frequency band. 

\vspace{+0.518mm}
\noindent\textbf{Spatially Variant Modulation.} 
After obtaining the output results for each frequency band, a modulation map $\mathbf{A}_b \in \mathbb{R}^{h \times w}$ is generated to control the influence of each frequency band at each spatial location. 
$\mathbf{A}_b$ can be easily implemented using a standard convolution layer followed by a sigmoid function. 
Consequently, the output feature map $\mathbf{Y}$ is computed as:
\vspace{-2mm}
\begin{equation}
\begin{aligned}
\mathbf{Y} = \sum_{b=0}^{B-1} (\mathbf{A}_b \odot \mathbf{Y}_b).
\end{aligned}
\vspace{-2mm}
\end{equation}
This approach allows FBM to adjust frequency responses dynamically at each spatial location, enhancing the model's ability to capture context-specific features across the image effectively.

\vspace{+0.518mm}
\noindent\textbf{Practical Implementation.} 
Mathematically, the obtation of $\mathbf{Y}_b$ in Equation~\eqref{eq:y_b} is equivalent to:
\begin{equation}
\begin{aligned}
\mathbf{Y}_b = \mathcal{F}^{-1}\left(\mathcal{F}(\mathbf{W}) \odot \left(\mathcal{M}_{b} \odot \mathcal{F}(\mathbf{X})\right)\right)
= \mathbf{W} * \mathbf{X}_b.
\end{aligned}
\end{equation}

The above derivation reveals that decomposing the convolution kernel into frequency bands (\ie, $\mathbf{W}_b= \mathcal{M}_{b} \odot \mathbf{W}$) is mathematically equivalent to decomposing the input feature map into corresponding frequency components (\ie, $\mathbf{X}_b = \mathcal{M}_{b} \odot \mathbf{X}$). This equivalence stems from the commutative property of convolution and the linearity of the Fourier transform. Specifically, modulating and convolving frequency-specific weights with the original feature map can be reinterpreted as first filtering the feature map into sub-bands and then convolving with the full kernel:
\vspace{-2mm}
\begin{equation}
\begin{aligned}
\mathbf{Y} = \sum_{b=0}^{B-1} \left(\mathbf{A}_b \odot \mathbf{X}_{b}\right) * \mathbf{W}.
\end{aligned}
\vspace{-2mm}
\end{equation}
This perspective bridges two seemingly distinct paradigms: frequency-adaptive weight decomposition and multi-band feature processing. The equivalent implementation not only circumvents the impracticality of infinite spatial support in ideal frequency filters but also provides implementation flexibility, one can choose to implement frequency decomposition on either weights or features based on computational constraints, while maintaining strict mathematical equivalence through Fourier duality.

\section{Experiment}

\noindent{\bf Datasets and Metrics.}
We evaluate our methods on challenging semantic segmentation datasets, including Cityscapes~\cite{cityscapes2016} and ADE20K~\cite{ade20k}, using mean Intersection over Union (mIoU) for segmentation~\cite{fcn2015, 2023casid, 2022levelAware, chen2022consistency, chen2024semantic} and Average Precision (AP) for object detection and instance segmentation~\cite{fasterRCNN2015, MaskRCNN2017}.

\vspace{+0.518mm}
\noindent{\bf Implementation Details.}
We follow the settings from the original papers for UPerNet~\cite{2018upernet}, Mask2Former~\cite{2022mask2former}, MaskDINO~\cite{2023pidnet}, Swin Transformer~\cite{2021swin}, and ConvNeXt~\cite{2022convnet, 2024kw}. On COCO~\cite{lin2014microsoft}, we adhere to standard practices~\cite{2022hornet, 2022dilatedatt, 2023internimage}, training detection and segmentation models for 12 epochs (1$\times$ schedule).
We empirically set the number of weights to $64$ for FDConv.
More details are described in the supplementary.

\begin{table}[t!]
\centering
\caption{
Results comparison on the COCO validation set~\cite{mscoco2014}. The numbers in brackets indicate the parameters of the backbone. Additionally, the notation $n \times$ denotes the convolutional parameter budget of each dynamic convolution relative to the standard convolution in the backbone.
}
\vspace{-2.18mm} 
\label{tab:ms-coco-results}
\scalebox{0.7628}{%
\begin{tabular}{l|r|r|c|c}
\toprule[1.28pt]
Models & Params & FLOPs &\footnotesize AP$^{box}$&\footnotesize AP$^{mask}$\\
\midrule
\textit{Faster R-CNN} & 43.80$_{(23.5)}$M & 207.1G & 37.2 &  \\
\midrule
+ CondConv{\color{gray}\tiny [NIPS2019]} (8$\times$)~\cite{2019condconv} & +90.0M & +0.01G & 38.1 & -  \\
+ DY-Conv{\color{gray}\tiny [ICLR2022]} (4$\times$)~\cite{2020dyconv} & +75.3M & +0.16G & 38.3 & - \\
+ DCD{\color{gray}\tiny [ICLR2021]}~\cite{2021DCD} & +4.3M & +0.13G & 38.1 &  \\
+ ODConv{\color{gray}\tiny [ICLR2022]} (4$\times$)~\cite{2022odconv} & +65.1M & +0.35G & 39.2 & -\\
\midrule
\rowcolor{gray!18}
+ FDConv (Ours) & +3.6M & +1.8G & \bf 39.4 &  \\
\midrule
\textit{Mask R-CNN} & 46.5$_{(23.5)}$M & 260.1 & 39.6 & 36.4 \\
\midrule
+ DY-Conv{\color{gray}\tiny [CVPR2020]} (4$\times$)~\cite{2020dyconv} & +75.3M & +0.16G & 39.6 & 36.6 \\
+ ODConv{\color{gray}\tiny [ICLR2022]} (4$\times$)~\cite{2022odconv} & +65.1M & +0.35G & 42.1 & 38.6 \\
+ KW{\color{gray}\tiny [ICML2024]} (1$\times$)~\cite{2024kw} & +2.5M & - & 41.8 & 38.4 \\
+ KW{\color{gray}\tiny [ICML2024]} (4$\times$)~\cite{2024kw} & +76.5M & - & 42.4 & \bf 38.9 \\
\midrule
\rowcolor{gray!18}
+ FDConv (Ours) & +3.6M & +1.8G & \bf 42.4 & 38.6 \\
\bottomrule[1.28pt]
\end{tabular}
}
\end{table}

\begin{table}[t!]
\color{black}
\caption{
\small 
Quantitative comparisons on semantic segmentation tasks with UPerNet~\cite{2018upernet} on the ADE20K validation set.
\vspace{-2.18mm}
}
\label{tab:ade20k}
\centering
\scalebox{0.828}{
\begin{tabular}{l|r|r|c|c}
\toprule[1.28pt]
\multirow{2}{*}{Method} & \multirow{2}{*}{Params} & \multirow{2}{*}{FLOPS} &\multicolumn{2}{c}{mIoU} \\
\cline{4-5}
& & & SS & MS \\
\midrule
ResNet-50~\cite{resnet2016} &66M &947G & 40.7 & 41.8 \\
ResNet-101~\cite{resnet2016} &85M &1029G & 42.9 & 44.0 \\
\midrule

R50 + PEDConv{\color{gray}\tiny [BMVC2021]}~\cite{2021PEConv} &72M &947G & 42.8 &  43.9 \\
R50 + ODConv{\color{gray}\tiny [ICLR2022]}($4\times$)~\cite{2022odconv} &131M &947G & 43.3 &  44.4 \\
R50 + KW{\color{gray}\tiny [ICML2024]}($4\times$)~\cite{2024kw} &141M &947G & 43.5 &  44.6 \\
\rowcolor{gray!18}
R50 + FDConv (Ours) & 70M &949G &\bf 43.8 & \bf 44.9 \\
\bottomrule[1.28pt]
\end{tabular}
}
\end{table}

\begin{table}[t!]
\centering
\caption{
Object detection and instance segmentation performance on the COCO dataset~\cite{mscoco2014} with the Mask R-CNN detector~\cite{MaskRCNN2017}.
All models are trained with a 1$\times$ schedule~\cite{2022dilatedatt, 2023internimage}.
\vspace{-2.18mm}
}
\scalebox{0.8588}{
\begin{tabular}{l|c|c|c|ccc|ccc|ccc|cccccc}
\toprule[1.28pt]
Model & Params & FLOPs &\footnotesize AP$^{box}$&\footnotesize AP$^{mask}$\\
\midrule
ConvNeXt-T~\cite{2022convnet} & 48M & 262G  & 43.4  & 39.7 \\
+ KW{\color{gray}\tiny [ICML2024]}~\cite{2024kw} & 52M & 262G &44.8 &40.6 \\
\rowcolor{gray!18}
+ FDConv (Ours) &51M &263G &\bf 45.2 &\bf 40.8 \\
\midrule
Swin-T\cite{2021swin} & 48M & 267G  & 42.7 & 39.3  \\
\rowcolor{gray!18}
+ FDConv (Ours) & 51M & 268G &\bf 44.5 &\bf 40.5 \\
\bottomrule[1.28pt]
\end{tabular}}
\label{tab:dinat}
\vspace{-2.18mm}
\end{table}

\begin{table}[tb!]
\centering
\caption{
Semantic segmentation results on Cityscapes~\cite{cityscapes2016} using the recent state-of-the-art Mask2Former~\cite{2022mask2former}. 
}
\vspace{-2.18mm}
\scalebox{0.8588}{
\begin{tabular}{l|l|ccccccc}
\toprule[1.28pt]
Model & Backbone & mIoU \\
\midrule
Mask2Former{\color{gray}\tiny [CVPR2022]}~\cite{2022mask2former} & ResNet-50 & 79.4 \\
\rowcolor{gray!18}
+ FDConv (Ours) & ResNet-50 &\bf 80.4 (+1.0) \\
\bottomrule[1.28pt]
\end{tabular}
}
\label{tab:mask2former_cityscapes}
\end{table}

\begin{table}[tb!]
\centering
\caption{
Semantic segmentation results with recent state-of-the-art large models Mask2Former~\cite{2022mask2former} and MaskDINO~\cite{2023maskdino} on ADE20K. Backbones pre-trained on ImageNet-22K are marked with $^\dagger$.
}
\vspace{-2.18mm}
\scalebox{0.8588}{
\begin{tabular}{l|l|ccccccc}
\toprule[1.28pt]
Model & Backbone & mIoU \\
\midrule
Mask2Former{\color{gray}\tiny [CVPR2022]}~\cite{2022mask2former} & Swin-B$^\dagger$ & 53.9 \\
\rowcolor{gray!18}
+ FDConv (Ours) & Swin-B$^\dagger$ &\bf 54.9 (+1.0) \\
\midrule
MaskDINO{\color{gray}\tiny [CVPR2023]}~\cite{2023maskdino} & Swin-L$^\dagger$ & 56.6  \\
\rowcolor{gray!18}
+ FDConv (Ours) & Swin-L$^\dagger$ &\bf 57.2 (+0.5)\\
\bottomrule[1.28pt]
\end{tabular}
}
\label{tab:mask2former}
\end{table}
\begin{figure}[t!]
\centering
\scalebox{0.9918}{
\begin{tabular}{cc}
\hspace{-2.918mm}
\includegraphics[width=0.998\linewidth]{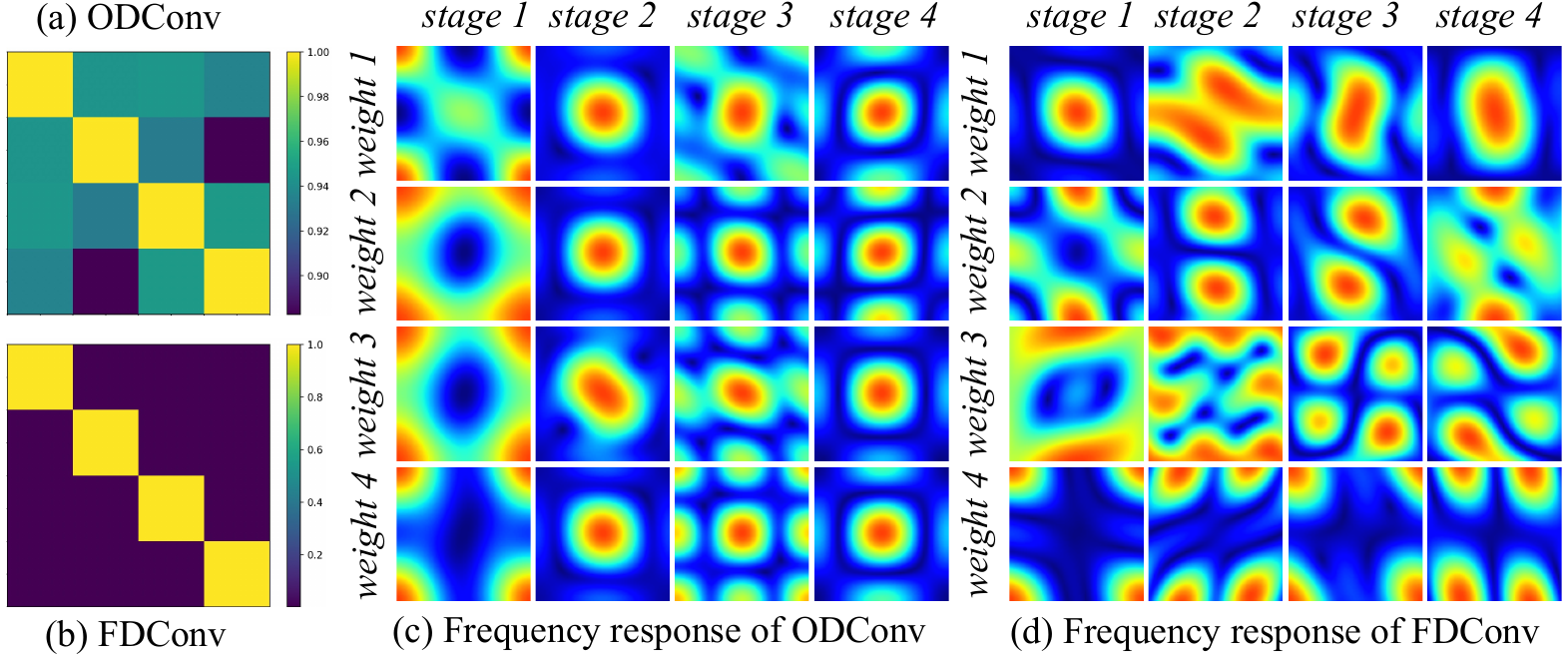} \\
\vspace{-6mm}
\end{tabular}}
\caption{
Weight similarity and frequency analyses.
(a) demonstrates that existing dynamic convolution methods, such as ODConv~\cite{2022odconv}, exhibit high cosine similarity ($>$0.88) among their 4 learned weights. 
The frequency analysis in (c) shows 4 representative ODConv layers from stage 1 to stage 4 of the model, and it demonstrates large homogeneity between the 4 weights. In contrast, the 4 weights of our proposed FDConv show zero similarity in (b), allowing each kernel to learn distinct and complementary features with diversified frequency response, as shown in (d).
}
\label{fig:similarity}
\vspace{-5mm} 
\end{figure}

\begin{figure*}[t!]
\centering
\scalebox{1.0}{
\begin{tabular}{cc}
\hspace{-3.918mm}
\includegraphics[width=1.0\linewidth]{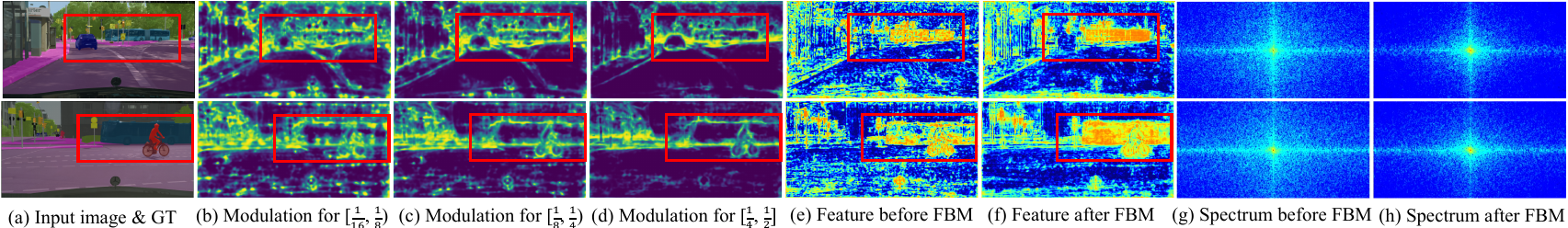} \\
\vspace{-6.58mm}
\end{tabular}}
\caption{
Visualization of Frequency Band Modulation. 
(a) shows the input images and their corresponding ground truth (GT). 
(b)–(d) display the modulation maps for different frequency bands, ranging from low to high. 
(e) and (f) visualize the feature frequency spectrum.
}
\label{fig:FBM_visualization}
\vspace{-2.18mm} 
\end{figure*}

\section{Main Results}
In this section, we evaluate our FDConv on a range of tasks, including object detection, instance segmentation, and semantic segmentation, using standard benchmarks such as COCO~\cite{mscoco2014}, ADE20K~\cite{ade20k}, and Cityscapes~\cite{cityscapes2016}. 

We compare our FDConv with state-of-the-art dynamic convolutional methods, including CondConv~\cite{2019condconv}, DY-Conv~\cite{2020dyconv}, DCD~\cite{2021DCD}, ODConv~\cite{2022odconv} and KW~\cite{2024kw}. The experiments demonstrate that FDConv not only achieves the highest performance across detection and segmentation tasks but also does so with large reduced parameter overhead.
Moreover, FDConv is highly versatile, it can easily combine with state-of-the-art ConvNet models like ConvNeXt~\cite{2022convnet} and apply to transformer architectures such as Swin-T~\cite{2021swin}, Mask2Former~\cite{2022mask2former}, and MaskDINO~\cite{2023maskdino}.
The experimental results demonstrate that FDConv achieves notable improvements over both conventional competitors and state-of-the-art baselines.

\vspace{+0.518mm}
\noindent\textbf{Object Detection.}
Table~\ref{tab:ms-coco-results} shows the results obtained by Faster R-CNN with various dynamic convolutional modules. Our FDConv module, despite adding only +3.6M parameters and +1.8G FLOPs, achieves an AP$^{box}$ of 39.4, 2.2\% improvement over the baseline and outperforms CondConv~\cite{2019condconv}, DY-Conv~\cite{2020dyconv}, and DCD~\cite{2021DCD}, and ODConv~\cite{2022odconv}, which require substantially higher parameter budgets. 
FDConv not only surpasses other methods in terms of accuracy but also achieves this with a minimal computational footprint, positioning it as a highly efficient enhancement for object detection tasks.

\vspace{+0.518mm}
\noindent\textbf{Instance Segmentation.} 
We further evaluate FDConv using Mask R-CNN~\cite{MaskRCNN2017} as the base model, following~\cite{2022odconv, 2024kw}. 
FDConv achieves an AP$^{box}$ of 42.4 and AP$^{mask}$ of 38.6, surpassing or matching recent high-performing methods such as ODConv~\cite{2022odconv} and KW~\cite{2024kw}.
Notably, while KW~\cite{2024kw} achieves marginally higher segmentation performance, it incurs a 4$\times$ increase in parameter cost (+76.5M), whereas FDConv adds only 3.6M.

\vspace{+0.518mm}
\noindent\textbf{Semantic Segmentation.}
As shown in Table~\ref{tab:ade20k}, FDConv achieves the highest mIoU scores, with a single-scale (SS) mIoU of 43.8. Notably, FDConv accomplishes this performance with fewer additional parameters (70M total) compared to ODConv~\cite{2022odconv} (131M) and KW~\cite{2024kw} (141M), underscoring its parameter efficiency while achieving superior segmentation quality.

\vspace{+0.518mm}
\noindent\textbf{Combination with Advanced Architectures.}
Additionally, we test FDConv on object detection and instance segmentation tasks using the COCO~\cite{mscoco2014} to examine its cross-architecture applicability. 
Table~\ref{tab:dinat} demonstrates that FDConv outperforms other methods, including KW~\cite{2024kw}, when applied to both ConvNeXt~\cite{2022convnet} and Swin Transformer~\cite{2021swin} backbones. It achieved an AP$^{\text{box}}$ of 45.2 with ConvNeXt-T~\cite{2022convnet} and 44.5 with Swin-T~\cite{2021swin}, along with enhanced AP$^{\text{mask}}$ scores. 
These results underscore FDConv's consistent generalization capabilities across various architectures.

\vspace{+0.518mm}
\noindent\textbf{Combination with Heavy Models.}
To assess the adaptability of our FDConv with advanced architectures, we incorporate FDConv into the state-of-the-art Mask2Former~\cite{2022mask2former} and MaskDINO~\cite{2023maskdino} frameworks. 
Table~\ref{tab:mask2former_cityscapes} shows that Mask2Former-ResNet-50 with FDConv achieves an mIoU improvement of +1.0 (from 79.4 to 80.4) on Cityscapes~\cite{cityscapes2016}.
On ADE20K~\cite{ade20k}, Table~\ref{tab:mask2former} highlights that with FDConv, Mask2Former-Swin-B~\cite{2022mask2former} achieves an mIoU improvement of +1.0 (from 53.9 to 54.9), while MaskDINO-Swin-L~\cite{2023maskdino} achieves an mIoU improvement of +0.5 (from 56.6 to 57.2). These consistent gains demonstrate that FDConv can effectively enhance heavy architectures.

\section{Analyses and Discussion}
We use ResNet-50~\cite{2017dilated} as the backbone model and conduct a comprehensive analysis of the proposed FDConv. 
Due to space limitations, more detailed analyses and the results of \underline{ablation studies} are provided in the \textit{supplementary material}.

\vspace{+0.518mm}
\noindent\textbf{Weight Similarity Analysis.} 
We analyze the diversity of learned features in FDConv by comparing weight similarity with existing dynamic convolution methods. As shown in Figure~\ref{fig:similarity}{\color{red}(a)}, traditional methods like ODConv~\cite{2022odconv} exhibit high cosine similarity ($\textgreater$ 0.88) among their learned weights, indicating significant redundancy. This redundancy limits the representational capacity of the model, as each kernel learns overlapping features.

In contrast, FDConv kernels exhibit zero cosine similarity (Figure~\ref{fig:similarity}{\color{red}(b)}), suggesting that each kernel captures unique, complementary features. This diversity enhances the model's expressiveness and adaptability.

\noindent\textbf{Weight Frequency Analysis.}
As shown in Figure~\ref{fig:similarity}{\color{red}(c)}, frequency analysis reveals that ODConv weights exhibit limited frequency diversity across different stages. In contrast, FDConv demonstrates a more diversified frequency response (Figure~\ref{fig:similarity}{\color{red}(d)}), capturing a broader range of frequency characteristics. This allows FDConv to model a richer set of features, further improving its ability to represent complex input data.

\vspace{+0.518mm}
\noindent\textbf{Feature Visualization for FBM.}
As shown in Figure~\ref{fig:FBM_visualization}, we visualize the modulation maps for each frequency band. For better performance, we empirically set the modulation map for the lowest frequency band to all 1.
We observe that higher modulation values are concentrated around object boundaries, with this effect becoming more pronounced in higher frequency bands (Figure~\ref{fig:FBM_visualization}{\color{red}(b)-(d)}). In contrast, lower frequency bands exhibit regions of high modulation within the objects themselves (Figure~\ref{fig:FBM_visualization}{\color{red}(c)}).

This selective modulation enables FDConv to suppress high frequencies in regions such as the background and object centers, which do not contribute significantly to accurate predictions. As seen in Figure~\ref{fig:FBM_visualization}{\color{red}(e)-(f)}, high-frequency noise in the feature map is largely reduced, and the spectrum in Figure~\ref{fig:FBM_visualization}{\color{red}(g)-(h)} further confirms the suppression of unnecessary high-frequency components.
Meanwhile, as shown in Figure~\ref{fig:FBM_visualization}{\color{red}(e)-(f)}, foreground features are enhanced, leading to more accurate and complete representations that benefit dense prediction tasks.

\section{Conclusion}
We introduced Frequency Dynamic Convolution (FDConv), which enhances the frequency adaptability of parallel weights without increasing parameter overhead. By incorporating Fourier Disjoint Weight (FDW), Kernel Spatial Modulation (KSM), and Frequency Band Modulation (FBM), FDConv addresses the limitations of existing dynamic convolution methods, including restricted frequency diversity in parallel weights and high parameter costs.

Our analysis shows that FDConv achieves greater frequency diversity, enabling better feature capture across spatial and frequency domains. 
Extensive experiments on object detection, segmentation, and classification demonstrate that FDConv outperforms prior state-of-the-art methods, with only a modest increase in parameter cost compared to others that incur much higher overhead.
FDConv can be easily integrated into existing architectures, including both ConvNets and vision transformers, making it a versatile and efficient solution for a wide range of computer vision tasks. 
We hope our analyses and finding would new direction for building more efficient and powerful vision models.

\section*{Acknowledgements}
This work was supported by the National Key R\&D Program of China (2022YFC3300704), the National Natural Science Foundation of China (62331006, 62171038, and 62088101), the Fundamental Research Funds for the Central Universities, and the JST Moonshot R\&D Grant Number JPMJMS2011, Japan.

{\small
\bibliographystyle{ieee_fullname}
\bibliography{./egbib.bib}
}
\end{document}

%% file: preamble.tex
\usepackage[dvipsnames]{xcolor}
